%% file: acl_latex.tex
\pdfoutput=1

\documentclass[11pt]{article}

\usepackage[svgnames]{xcolor}

\usepackage[final]{acl}

\usepackage{times}
\usepackage{latexsym}

\usepackage[T1]{fontenc}

\usepackage[utf8]{inputenc}

\usepackage{microtype}

\usepackage{inconsolata}

\usepackage{graphicx}

\usepackage{subfig}
\usepackage{pgfplots}
\usepgfplotslibrary{fillbetween}
\usetikzlibrary{patterns}
\usetikzlibrary{pgfplots.groupplots}
\pgfplotsset{compat=1.17}
\usepackage{lscape}
\usepackage{multirow, makecell}
\usepackage{dblfloatfix}
\usepackage{footnote}
\usepackage{graphicx}
\usepackage{tikz}
\usepackage{booktabs, tabularx}
\usepackage{amssymb}
\usepackage{amsmath}
\usepackage{colortbl}
\usepackage{soul}
\usepackage{mathtools}

\newcommand{\ctext}[3][RGB]{%
  \begingroup
  \definecolor{hlcolor}{#1}{#2}\sethlcolor{hlcolor}%
  \hl{#3}%
  \endgroup
}

\newcommand{\evidence}[1]{\ctext[RGB]{248,245,175}{#1}}
\newcommand{\head}[1]{\textsl{\ctext[RGB]{187,234,230}{#1}}}
\newcommand{\tail}[1]{\textsl{\ctext[RGB]{191,248,173}{#1}}}
\newcommand{\poison}[1]{\textsl{\ctext[RGB]{254,183,179}{#1}}}

\usepackage{amssymb}
\usepackage{tikz}
\usepackage{cancel}
\newcommand{\mathboldface}[1]{\boldsymbol{#1}}
\newcommand{\bm}[1]{\mathboldface{#1}}

\definecolor{coldercolor}{HTML}{00A6ED}

\title{
Collapse of Dense Retrievers: \\Short, Early, and Literal Biases Outranking Factual Evidence
}

\author{
    Mohsen Fayyaz$^{1}$ ~ Ali Modarressi$^{2,3}$ ~
    \textbf{Hinrich Schütze$^{2,3}$}  ~ \textbf{Nanyun Peng$^{1}$} \\
    $^1$ University of California, Los Angeles \\ 
    $^2$ CIS, LMU Munich ~ $^3$ Munich Center for Machine Learning \\
    \texttt{mohsenfayyaz@cs.ucla.edu} ~
    \texttt{amodaresi@cis.lmu.de} ~
    \texttt{violetpeng@cs.ucla.edu}
}

\begin{document}
\maketitle
\begin{abstract}
Dense retrieval models are commonly used in Information Retrieval (IR) applications, such as Retrieval-Augmented Generation (RAG). Since they often serve as the first step in these systems, their robustness is critical to avoid downstream failures.
In this work, we repurpose a relation extraction dataset (e.g., Re-DocRED) to design controlled experiments that quantify the impact of heuristic biases, such as a preference for shorter documents, on retrievers like Dragon+ and Contriever.
We uncover major vulnerabilities, showing \textit{retrievers favor shorter documents, early positions, repeated entities, and literal matches}, all while ignoring the answer's presence!
Notably, when multiple biases combine, models exhibit catastrophic performance degradation, selecting the answer-containing document in less than 10\% of cases over a synthetic biased document without the answer. 
Furthermore, we show that \textit{these biases have direct consequences for downstream applications like RAG}, where retrieval-preferred documents can mislead LLMs, resulting in a 34\% performance drop than providing no documents at all. (\href{https://huggingface.co/datasets/mohsenfayyaz/ColDeR}{Dataset \& Leaderboard})\footnote{\textcolor{coldercolor}{\Large{$\bm{\ast}$}}\,Code and benchmark dataset are available at \href{https://huggingface.co/datasets/mohsenfayyaz/ColDeR}{https://huggingface.co/datasets/mohsenfayyaz/ColDeR}.}

\end{abstract}

\section{Introduction}

\input{figs/radar_plot}

Retrieval-based language models have demonstrated strong performance on a range of knowledge-intensive NLP tasks \cite{RAGLewis2020, asai-etal-2023-retrieval, gao2024retrievalaugmentedgenerationlargelanguage}. At the core of these models is a retriever that identifies relevant context to ground the generated output. Dense retrieval methods such as Contriever \citep{izacard2021contriever}—where passages or documents are stored as learned embeddings—are especially appealing for their scalability across large knowledge bases and handling lexical gaps \cite{ni2022large, ScalingShao2024}, compared to alternatives like BM25 \cite{INR-019-BM25} or ColBERT\cite{10.1145/3397271.3401075-colbert}. Despite their widespread use, relatively little is understood about how these dense models encode and organize information, leaving key questions about their robustness against adversarial attacks unanswered.

\footnotetext{$\, p<.05 \Rightarrow |t| \geq 1.97, df = 249$}

Existing evaluations often focus on downstream task performance, as seen in benchmarks like BEIR \cite{thakur2021beir}, without probing the underlying behaviors of retrievers. Some studies have analyzed specific issues in information retrieval (IR) models, such as position bias \cite{coelho-etal-2024-dwell} or lexical overlap \citep{ram-etal-2023-token}. 

\input{tables/examples}

In this work, we study multiple biases' impact on retrievers—both individually and in combination—for the first time. To enable fine-grained control over document structure and factual positioning, we repurpose a document-level relation extraction dataset (Re-DocRED \cite{tan-etal-2022-revisiting}).

We first investigate biases individually, identifying tendencies such as an \textbf{over-prioritization of document beginnings, document brevity, repetition of matching entities, and literal matches at the expense of ignoring answer presence.} Our statistical approach, illustrated in Figure~\ref{fig:radar_plot}, allows for comparative analysis across different biases. Additionally, we explore the interplay between these biases and propose an adversarial benchmark that combines multiple vulnerabilities.

We further study combining multiple biases and reveal concerning patterns in current retriever architectures. \textbf{When exposed to multiple interacting biases, even top-performing models exhibit dramatic degradation}, selecting the answer-containing document over the foil document—filled with biases—less than 10\% of the time. Moreover, we demonstrate that \textbf{these biases can be exploited to manipulate Retrieval-Augmented Generation (RAG)}, causing retrievers to favor misleading or adversarially constructed documents, which misguide LLMs into using incorrect information and ultimately degrade its performance.


\section{Related Work}
\paragraph{Benchmarking in Information Retrieval}
Popular benchmarks like BEIR \citep{thakur2021beir, guo-etal-2021-multireqa, petroni-etal-2021-kilt, muennighoff-etal-2023-mteb} have played a crucial role in evaluating retrieval models across diverse datasets and tasks.
In addition to general IR benchmarks, domain-specific benchmarks such as COIR \citep{li2024coircomprehensivebenchmarkcode} for code retrieval and LitSearch \citep{ajith-etal-2024-litsearch} for scientific literature search address retrieval challenges in specialized domains.
While these benchmarks have advanced the evaluation of IR models, they primarily focus on downstream performance rather than conducting systematic analyses of biases inherent in retrieval systems.

\paragraph{Information Retrieval Model Analysis}
Prior work in information retrieval has explored various dimensions of retrieval performance, including positional biases \citep{coelho-etal-2024-dwell}. Studies have also examined how dense retrievers exhibit biases towards common entities and struggle with OOD scenarios \citep{sciavolino-etal-2021-simple}. Furthermore, analysis by projecting representations to the vocabulary space has shown that supervised dense retrievers tend to learn relying heavily on lexical overlap during training \citep{ram-etal-2023-token}. Similarly, \citet{behnamghader-etal-2023-retriever} has indicated that Dense Passage Retrieval (DPR) models often fail to retrieve statements requiring reasoning beyond surface-level similarity.
Furthermore, neural IR models have been shown to exhibit over-penalization for extra information, where adding a relevant sentence to a document can unexpectedly decrease its ranking \citep{Usuha-etal-2024-Over-penalization-for-Extra-Information}.
Additionally, \citet{reichman-heck-2024-dense} takes a mechanistic approach to analyze the impact of DPR fine-tuning, showing that while fine-tuned models gain better access to pre-trained knowledge, their retrieval capabilities remain constrained by the pre-existing knowledge in their base models.
Further, \citet{macavaney-etal-2022-abnirml} provides a framework for analyzing neural IR models, identifying key biases and sensitivities in these models. 

\paragraph{Adversarial Attacks in Information Retrieval}

Numerous studies have explored various dimensions of robustness in information retrieval, including aspects related to adversarial robustness \citep{10.1145/3626772.3661380}.
Adversarial perturbations, for instance, have been shown to significantly degrade BERT-based rankers' performance, revealing their brittleness to subtle modifications \citep{10.1145/3539813.3545122}. 
Existing retrieval attack methods primarily encompass corpus poisoning \citep{10646819, zhong-etal-2023-poisoning}, backdoor attacks \citep{long2024whispersgrammarsinjectingcovert}, and encoding attacks \citep{10.1145/3607199.3607220}. 

While previous work has analyzed some retrieval biases, most studies focus on task-specific supervised models and a single aspect in isolation. Our work provides a comprehensive comparative analysis of popular retrieval models across multiple dimensions of vulnerability. We systematically investigate how these biases interact and affect the retrieval capabilities of dense retrievers. By repurposing a relation extraction dataset, we gain precise control over factual information in documents, enabling a rigorous evaluation of retrieval robustness. This multi-dimensional approach provides a nuanced understanding of the strengths and weaknesses of dense retrievers.

\section{Experiments}
\input{figs/decompx_heatmap}

\subsection{A Framework for Identifying Biases in Retrievers}
To gain fine-grained control over the facts present in a document, we take a novel approach by repurposing a relation extraction dataset that provides relation-level fact granularity. This enables a structured analysis of retrieval biases by explicitly linking queries to individual factual statements.

One such dataset is DocRED \cite{yao-etal-2019-docred}, a relation extraction dataset constructed from Wikipedia and Wikidata. DocRED consists of human-annotated triplets (\head{head entity}, relation, \tail{tail entity})—for example, (Albert Einstein, educated at, University of Zurich). However, DocRED suffers from a significant false negative issue, as many valid relations are missing from the annotations.
To address this, we use \mbox{Re-DocRED} \cite{tan-etal-2022-revisiting}, a re-annotated version of DocRED that recovers missing facts, leading to more complete and reliable annotations. 

To construct a retrieval dataset from Re-DocRED, we map each relation to a query template (Templates are in Table~\ref{tab:query_templates}). For example, for the relation "educated at," we use the template "Where was \{Head Entity\} educated?" This transformation allows us to systematically examine how retrievers handle different types of factual queries.


\input{tables/models_performance_nq}

The answers to these queries are the tail entities found in the evidence sentences provided by the dataset. For our analysis, we ensure that each query has a single evidence sentence ($S_{ev}$) within the original document ($S_{ev} \in D_{orig}$) that contains both the head and tail entities. This constraint makes the sentence self-contained, allowing for precise control over the document structure in subsequent sections.
We also introduce the notation $S^{+h}_{-t}$ for sentences in $D_{orig}$ that contain the head entity but not the tail entity, and $S^{-h}_{-t}$ for sentences that do not contain either entity. In each of the following sections, we will use this notation to construct a pair of document sets, $D_1$ and $D_2$, enabling a systematic investigation of retrieval score variations and potential biases.
As a result, for each of our six analysis settings, we compile 250 queries, each with a single corresponding gold document, based on the test and validation sets of Re-DocRED.

\subsection{Models Performance \& Bias Discovery}

First, we evaluate several dense retrievers on the NQ dataset \citep{kwiatkowski-etal-2019-natural}, comparing their performance using nDCG@10 and Recall@10 metrics. Table~\ref{tab:models_performance_nq} shows that Dragon models lead in performance, and the significant improvement of fine-tuned Contriever (Contriever MSMARCO) over its unsupervised counterpart highlights the importance of supervision and task-specific adaptation. Models also differ in pooling mechanisms, with Contriever using average pooling and others using CLS pooling. For details, refer to the appendix~\ref{sec:models_performance}.

In our preliminary analysis, we utilized DecompX \cite{modarressi-etal-2023-decompx, modarressi-etal-2022-globenc}, a method that decomposes the representations of encoder-based models such as BERT into their constituent token-based representations. By applying DecompX to the embeddings generated by dense retrievers, we obtain decomposed representations for both the query and the document. Instead of using the original embeddings, we compute the similarity score via a dot product of the decomposed vectors. This approach enables us to visualize the contribution of each query and document token to the final similarity score as a heatmap (Figure~\ref{fig:decompx_heatmap}), revealing biases in token-level interactions.

In our preliminary error analysis of 60 retrieval failure examples, we identified potential biases and limitations in the models (Table~\ref{tab:preliminary_analysis}). Figure~\ref{fig:decompx_heatmap} highlights some of these biases, such as Literal Bias, where the term "esteban gomez" fails to match "estevao gomez," reflecting a preference for exact matching. In subsequent sections, we design experiments and perform statistical tests to evaluate these observed biases.

\subsection{Bias Types in Dense Retrieval}
\label{sec:bias_types}
The following experiments are meticulously designed to control for all other factors and biases, isolating the specific bias under evaluation.

\subsubsection{Answer Importance}

An effective retrieval model should accurately identify the query's intent. It should retrieve relevant documents that address the query, rather than just matching entities.
To assess whether dense retrieval models truly recognize the presence of answers or merely focus on entity matching, we developed a controlled experimental framework.
Our experimental design contrasts two carefully constructed document types. 
\textbf{1. Document with Evidence}:  Contains a leading evidence sentence with both the head entity and the tail entity (answer).
\textbf{2. Document without Evidence $\bm{D_2}$:} Contains a leading sentence with only the head entity but no tail. 
\begin{equation}
\small
\begin{aligned}
    \bm{D_1}& \coloneq \bm{S_{ev}}+\sum_{S^{-h}_{-t} \in D_{orig}} S^{-h}_{-t} \\
     \bm{D_2}& \coloneq S^{+h}_{-t} + \sum_{S^{-h}_{-t} \in D_{orig}} S^{-h}_{-t}
\end{aligned}
\end{equation}
Here, $S^{+h}_{-t}$ is another sentence from $D_{orig}$ that replaces the original evidence $S_{ev}$ while containing the head entity but not containing the tail entity to isolate the impact of answer presence.
The remainder of both documents consists of neutral sentences $S^{-h}_{-t} \in D_{orig}$, carefully filtered to exclude any sentences containing similar head relations or tail entities. This ensures the answer information appears exclusively in the leading sentence of the evidence document. We strategically positioned the key sentences at the beginning of both documents to mitigate potential position bias effects, which we analyze in subsequent sections.
An example of this setup is presented in Table~\ref{tab:examples} (Answer Impact).

\input{figs/tail_ttest}

To quantify the models' ability to distinguish between these document types, we employ \mbox{\textbf{Paired t-Test}}\footnote{Using ttest\_rel function of SciPy \cite{2020SciPy-NMeth}.} to analyze the difference in similarity scores. The t-test statistic (t) is calculated as:

\begin{equation}
    t = \frac{\bar{d}}{SE(d)} = \frac{\text{Average Difference}}{\text{Standard Error}}
\end{equation}

where $\bar{d}=mean\left(M(Q, D_1) - M(Q, D_2)\right)$ is the mean difference between paired observations\footnote{$M$ is the retriever model's score}, and $SE(d)$ is the standard error of these differences\footnote{$SE=\frac{\sigma}{\sqrt{n}}$}. A positive t-statistic indicates higher scores for $D_1$ documents, while negative values suggest a preference for $D_2$ documents. In this scenario, positive values are desirable as they indicate the model prefers $D_1$, which contains the answer, over $D_2$, which does not.

As shown in Figure~\ref{fig:tail_ttest}, our analysis reveals variations across models. Dragon+ and Dragon-RoBERTa demonstrate superior tail recognition, achieving the highest positive t-statistics. In contrast, Contriever exhibits poor performance, yielding negative t-statistics that indicate a failure to properly distinguish answer-containing passages.

The vanilla Contriever's underwhelming performance can be attributed to its unsupervised training methodology, which differs from models trained on MS MARCO \cite{bajaj2018msmarcohumangenerated}. While MS MARCO provides supervised training with explicit query-passage relevance labels, Contriever employs unsupervised contrastive learning. It generates positive pairs through data augmentation from document segments and derives negative examples implicitly via in-batch sampling from other texts. This training approach, while efficient for general text representation, appears insufficient for developing the fine-grained discrimination needed to understand query intent in retrieval tasks.

\subsubsection{Position Bias}
\input{figs/position_evidence_dot}

Position bias refers to the preference of retrieval models for information located in specific positions within a document, typically favoring content at the beginning over content appearing later. This bias is problematic as it may lead to the underrepresentation of relevant information that is positioned deeper within documents, thus reducing the overall retrieval quality and fairness.

Our analysis reveals a strong positional bias in dense retrievers, with models consistently prioritizing information at the beginning of documents. As shown in Figure~\ref{fig:position_evidence_dot}, we conducted paired t-tests comparing retrieval scores when the evidence sentence is placed at different positions to scores when it is placed at the document’s beginning ($M(Q, D_i)-M(Q, D_1)$).

\begin{equation}
\small
\begin{aligned}
    \bm{D_1}& \coloneq \bm{S_{ev}}+{}^{1}S^{-h}_{-t}+{}^{2}S^{-h}_{-t}+{}^{3}S^{-h}_{-t}+...+{}^{n}S^{-h}_{-t} \\
     \bm{D_2}& \coloneq {}^{1}S^{-h}_{-t}+ \bm{S_{ev}} + {}^{2}S^{-h}_{-t} + {}^{3}S^{-h}_{-t} + ... + {}^{n}S^{-h}_{-t}\\
     \bm{D_3} & \coloneq {}^{1}S^{-h}_{-t} + {}^{2}S^{-h}_{-t} + \bm{S_{ev}} + {}^{3}S^{-h}_{-t} + ... + {}^{n}S^{-h}_{-t}
\end{aligned}
\end{equation}

To ensure fairness, the examples were curated so that the remaining content was free of any evidence or head entity ($S^{-h}_{-t}$) like the last section. This design ensured that the evidence's position was the sole factor under evaluation. The consistently negative t-statistics across models in Figure~\ref{fig:position_evidence_dot} confirm a strong bias favoring content at document beginnings.\footnote{Fig.~\ref{fig:radar_plot} shows the impact of evidence placement (beginning vs. end), detailed in Appendix~\ref{sec:appendix}, with an example in Table~\ref{tab:examples}.}
This bias is most pronounced in Dragon-RoBERTa and Contriever-MSMARCO, which show the most negative t-statistics, indicating severe degradation in recognizing evidence further into the document. While Dragon+ and RetroMAE perform better, their negative t-statistics still confirm position bias in these models.

These findings align with recent research by \citet{coelho-etal-2024-dwell}, who demonstrated that positional biases emerge during the contrastive pre-training phase and worsened through fine-tuning on MS MARCO dataset with T5 \cite{raffel2020exploring-t5} and RepLLaMA \cite{rankllama} models. This can significantly impact retrieval performance when relevant information appears later in documents.

\subsubsection{Literal Bias}
Retrievers should ideally recognize semantic equivalence across different surface forms of the same entity. For instance, a robust model should understand that "Gomes" and "Gomez" refer to the same person, or that "US" and "United States" represent the same entity. However, our analysis reveals that current models exhibit a strong bias toward exact literal matches rather than semantic matching.

\input{tables/literal_bias}

In our dataset, each head entity can be represented by multiple alternative names. To investigate literal bias, we created different combinations of query and document by replacing all head entities with the shortest or longest name variants as illustrated in Table~\ref{tab:examples} (Literal Bias). For example, an entity might be represented as "NYC" (shortest) or "New York City" (longest), allowing us to test how the model performs when matching different combinations of these representations.

Table~\ref{tab:literal_bias} presents the paired t-test statistics comparing different combinations of name selections in queries and documents. The results consistently show positive statistics when Query 1 and Document 1 contain similar name representations. 
For our subsequent analysis of bias interplay, we specifically examine the comparison between two scenarios (Figure~\ref{fig:literal_ttest}): one where both query and document use the shortest name variant (short-short) versus cases where the query uses the short name but the document contains the long name variant (short-long). This corresponds to +14.37 and +16.62 in Table~\ref{tab:literal_bias} for Contriever and Dragon+, respectively.\footnote{We avoid long-long combinations to control for confounding effects, as they span multiple tokens and may introduce repetition bias due to token overlap}


\subsubsection{Brevity Bias}

Brevity bias refers to the tendency to favor concise text, such as a single evidence sentence, over longer documents that include the same evidence alongside additional context. 
This bias is problematic because retrievers may favor a shorter, non-relevant document over a relevant one. We will discuss this potential hazard further in \S\ref{Sec:RAG}.

Here, we performed paired t-tests to compare the similarity scores of queries with two sets of documents: \mbox{\textbf{(1) \textit{Single Evidence}}}, consisting of only the evidence sentence, and \mbox{\textbf{(2) \textit{Evidence+Document}}}, consisting of the evidence sentence followed by the rest of the document. The examples are carefully selected to ensure the evidence sentence includes both the head and tail entity and the rest of the document contains no repetition of the head entity or additional evidence.

\begin{equation}
\small
\begin{aligned}
    \bm{D_1}& \coloneq \bm{S_{ev}} \\
     \bm{D_2}& \coloneq \bm{S_{ev}} + \sum_{S^{-h}_{-t} \in D_{orig}} S^{-h}_{-t}
\end{aligned}
\end{equation}

Figure~\ref{fig:radar_plot} and \ref{fig:brevity_ttest}, illustrate the paired t-test statistics, where significant positive values indicate a strong bias toward brevity, as models assign higher scores to concise texts ($D_1$) than to longer ones with the same evidence ($D_2$).
This behavior likely stems from the way dense passage retrievers compress document representations. Most retrievers use either a mean-pooling strategy or a [CLS] token-based method. Both methods struggle with integrating useful evidence into the representation when unrelated content is present, leading to a ``pollution effect.'' As a result, the additional context in longer documents dilutes the importance of the evidence, causing retrievers to favor concise input. 


\subsubsection{Repetition Bias}

Repetition bias refers to the tendency of retrieval models to prioritize documents or passages with repetitive content, particularly repeated mentions of head entities present in the query. This bias is problematic as it may skew retrieval results toward redundant or verbose documents, undermining the goal of surfacing concise and diverse information.

\input{figs/repeatition_bias_heatmap_contriever}

To analyze repetition bias, we conducted an experiment evaluating the average retrieval dot product score of the models for samples with varying document lengths and head entity repetitions (Figure~\ref{fig:repetition_bias_heatmap_contriever} and \ref{fig:repetition_bias_heatmap}). 
A key concern is that longer documents naturally have a higher chance of lexical overlap with the query, as they may contain more repeated mentions of the head entity. This makes it difficult to disentangle the effects of document length from the number of entity repetitions. Therefore, we structure our analysis to separately examine these two factors. 
Our findings (Figure~\ref{fig:repetition_bias_heatmap_contriever}) reveal that the retrieval score increases with the number of head entity mentions, indicating a preference for documents with repeated entities. Conversely, the retrieval score decreases as document length increases, suggesting that longer documents are penalized despite potential relevance. Figure~\ref{fig:repetition_bias_heatmap} in the appendix generalizes these observations across all models. This experiment highlights the trade-off between repetition and document length, emphasizing the need for retrieval systems to balance these factors to mitigate bias.

We further explored this phenomenon through the results shown in Figures \ref{fig:radar_plot} and \ref{fig:repetition_ttest}. Here, we performed paired t-tests to compare the dot product similarity scores of queries with two sets of documents: \textbf{(1) \textit{More Heads}}, comprising an evidence sentence and two sentences containing head mentions but no tails, and \textbf{(2) \textit{Fewer Heads}}, comprising an evidence sentence and two sentences without head or tail mentions from the document (Table~\ref{tab:examples}). 

\begin{equation}
\small
\begin{aligned}
    \bm{D_1}& \coloneq \bm{S_{ev}}+S^{+h}_{-t}+S^{+h}_{-t} \\
     \bm{D_2}& \coloneq \bm{S_{ev}}+S^{-h}_{-t}+S^{-h}_{-t}
\end{aligned}
\end{equation}

Positive paired t-test values indicate higher similarity for sentences with more head mentions (Figure~\ref{fig:repetition_ttest}). The results strongly suggest that the model favors sentences with repeated heads, confirming the presence of repetition bias.

\subsection{Interplay Between Bias Types}
\label{sec:interplay}


To understand how different biases interact and amplify retrieval model weaknesses, we conduct a systematic analysis using a controlled 250-sample dataset across all experiments. This consistent sample size ensures comparability of paired t-test statistics across bias types and provides a robust basis for evaluating their interplay.

As illustrated in Figure~\ref{fig:radar_plot}, the paired t-test results reveal that brevity bias, literal bias, and position bias are the most problematic for dense retrievers. 
In contrast, repetition bias, while still detrimental, exhibits a relatively lower impact, suggesting that models are slightly more robust against this type of bias. Answer importance demonstrates an acceptable distinction between evidence-containing and no-evidence documents. However, the scores are not as strong as one would expect from models designed for accurate answer retrieval, highlighting the need for further improvement in this area.

\input{tables/foil_acc}

To further investigate the compounded effects of multiple biases, we conducted another experiment that combines several bias types into a single challenging setup. 
In this experiment, we created two document types.
\textbf{1) Foil Document with Multiple Biases:} This document contains multiple biases, such as repetition and position biases. It includes two repeated mentions of the head entity in the opening sentence, followed by a sentence that mentions the head but not the tail (answer). So it does not include the evidence.
\textbf{2) Evidence Document with Unrelated Content:} This document includes four unrelated sentences from another document, followed by the evidence sentence with both the head and tail entities. The document ends with the same four unrelated sentences. An example is shown in Table~\ref{tab:examples} (Foil vs. Evide.).\footnote{$\tilde{S}$ are sentences from an unrelated document}

\begin{equation}
\small
\begin{aligned}
    \bm{D_1}& \coloneq 2\times h+S^{+h}_{-t} \\
     \bm{D_2}& \coloneq 4\times \tilde{S}^{-h}_{-t}+\bm{S_{ev}}+4\times \tilde{S}^{-h}_{-t}
\end{aligned}
\end{equation}

Table~\ref{tab:foil_acc} presents the accuracy (proportion of times the model prefers $D_2$ over $D_1$), paired t-test statistics, and p-values.
The results are striking: all models exhibit extremely poor performance, with accuracy dropping below 10\%. The paired t-test statistics are highly negative across all models, indicating a consistent preference for foil documents over the correct evidence-containing ones. This outcome highlights the severity of bias interplay and its detrimental impact on model reliability. Furthermore, a sufficient number of biased documents can potentially cause the model to select all top-k documents from only biased results.



\subsection{Impact on RAG}
\label{Sec:RAG}
\input{tables/rag}
To assess the impact of the identified vulnerabilities on RAG systems, we use GPT-4o models \citep{openai2024gpt4ocard} and provide them with different versions of the reference document for a given query. Additionally, we construct a poisoned document by modifying the foil document from \S\ref{sec:interplay}, introducing a poisoned evidence sentence (Table~\ref{tab:examples_poison}). Specifically, we generate this sentence using GPT-4o by replacing the tail entity with a contextually plausible but entirely incorrect entity. This approach ensures that the poisoned document both exploits the previously discussed retrieval biases and contains an incorrect answer to the query.\footnote{Despite this, retrievers prefer the poisoned document over the evidence document in 100\% of cases (Table~\ref{tab:poison_acc}).}

\begin{equation}
\small
\begin{aligned}
    \bm{D_1}& \coloneq 2\times h+S^{+h}_{-t} + S^{+h}_{+PoisonTail}\\
     \bm{D_2}& \coloneq 4\times \tilde{S}^{-h}_{-t}+\bm{S_{ev}}+4\times \tilde{S}^{-h}_{-t}
\end{aligned}
\end{equation}


Table~\ref{tab:rag} reports the RAG accuracy,\footnote{Evaluated using GPT-4o. Prompts in Table~\ref{tab:rag_prompts}} showing that, as expected, providing the evidence document enables the LLM to achieve high accuracy. However, since retrievers prefer the foil document from \S\ref{sec:interplay}, which lacks evidence, LLM performance drops to levels near\footnote{Slightly lower, as the model sometimes abstains by stating, ``The document does not provide information.''} the no-document condition. This preference is concerning, as it allows biases to be exploited, making certain documents more likely to be retrieved despite embedding incorrect information. This is evident with the poisoned document, which degrades performance even worse than presenting no document by introducing false facts.
In summary, \textbf{retriever biases can mislead RAG systems by providing poisoned or non-informative documents}, ultimately harming performance.

\section{Conclusions}
In this work, we introduced a comprehensive framework for analyzing biases in dense retrieval models. By leveraging a relation extraction dataset (Re-DocRED), we constructed a diverse set of controlled experiments to isolate and evaluate specific biases, including literal, position, repetition, and brevity biases as well as the answer's importance. 

Our findings reveal that retrieval models often prioritize superficial patterns, such as exact string matches, repetitive content, or information positioned early in documents, over deeper semantic understanding and the existence of the answer. Moreover, when multiple biases combine, retriever performance deteriorates dramatically. 

Furthermore, our analysis shows that retriever biases can undermine RAG’s reliability by favoring poisoned or non-informative documents over evidence-containing ones, leading to degraded performance of LLMs. These findings underscore the need for dense retrieval models that are robust to biases and capable of prioritizing semantic relevance. 



\section*{Limitations}
\paragraph{Quality of the Relation Extraction Dataset}
Our framework relies on a relation extraction dataset, making both annotation accuracy (precision) and completeness (recall) critical. We use Re-DocRED, which addresses annotation issues in DocRED, but it may still contain imperfections that introduce minor noise into our experiments. To mitigate this, we employ statistical tests and report error margins and p-values to ensure the robustness of our findings.

\paragraph{Limitations of RAG Evaluation by LLMs}
In our RAG experiments, we utilized GPT-4o models and carefully designed prompts (Table~\ref{tab:rag_prompts}) to poison documents, generate answers using RAG, and evaluate the results against gold-standard answers. Although GPT-4o is one of the most advanced models available, it is not infallible and may introduce some variance in the RAG results and evaluations. Nevertheless, we believe the observed trends and findings remain valid given the model's high performance and the consistency of our experimental setup.

\section*{Acknowledgements}

This research is partly supported by a National Science Foundation CAREER award \#2339766 and an Amazon AGI Research Award.
We thank Jia-Chen Gu for valuable discussions and feedback. We also appreciate the insights from our peers, and we are grateful to the anonymous reviewers for their constructive comments.


\newpage


\bibliography{custom}

\appendix
\counterwithin{figure}{section}
\counterwithin{table}{section}

\newpage

\section{Appendix}
\label{sec:appendix}

\subsection{More Models}
\label{sec:more_models}
Our study focuses on dense retrievers that generate a single embedding per document, ensuring efficiency. In contrast, newer models like ColBERT \cite{10.1145/3397271.3401075-colbert} prioritize accuracy at the cost of efficiency. Specifically, ColBERT employs late interaction, requiring per-token embeddings for both queries and documents, which increases computational and storage demands, especially problematic for long documents. Despite these trade-offs, we evaluated ColBERT and ReasonIR-8B \cite{shao2025reasonirtrainingretrieversreasoning} on the Foil dataset, and their performance results are in Table~\ref{tab:foil_acc}. Despite their higher computational cost, these models still fail on the Foil dataset, with under 9\% correct document preferences.

\subsection{Models Downstream Performance}
\label{sec:models_performance}
We evaluate several dense retrievers on the Natural Questions (NQ) dataset \citep{kwiatkowski-etal-2019-natural}, comparing their performance using standard retrieval metrics: nDCG@10 and Recall@10\footnote{Using BEIR framework \citep{thakur2021beir}}. The models differ in training objectives, datasets, and pooling mechanisms, offering a comprehensive view of their retrieval capabilities in our experimental setup. Table~\ref{tab:models_performance_nq} (and \ref{tab:models_performance}) summarizes the results.





Dragon RoBERTa and Dragon+ \cite{lin-etal-2023-train} demonstrate the highest performances due to diverse data augmentations and multiple supervision sources, which progressively enhance their generalization.\footnote{Dragon RoBERTa is initialized from RoBERTa and Dragon+ from RetroMAE}

COCO-DR \cite{yu-etal-2022-coco} adopts continuous contrastive learning and implicit distributionally robust optimization (DRO) to address distribution shifts in dense retrieval tasks. It exhibits moderate performance, scoring lower than Dragon models.

Contriever \cite{izacard2021contriever} uses unsupervised contrastive learning but performs poorly without fine-tuning (nDCG@10: 0.25). Fine-tuning on MSMARCO significantly improves its performance (nDCG@10: 0.50), underscoring the importance of fine-tuning for robust retrieval.

RetroMAE \cite{xiao-etal-2022-retromae}, which introduces a retrieval-oriented pre-training paradigm based on Masked Auto-Encoder (MAE), featuring innovative designs like asymmetric masking, achieves slightly lower performance (nDCG@10: 0.48) compared to fine-tuned Contriever.

\input{tables/foil_acc_more}
\input{tables/models}
\input{tables/models_performance}
\input{tables/preliminary_analysis}

The models also differ in their pooling mechanisms. Contriever uses average pooling, where token representations are averaged to form a dense vector for retrieval. In contrast, the other models use CLS pooling, where the representation of the [CLS] token is taken as the sentence embedding.

In summary, Dragon models lead in performance, and the significant improvement of fine-tuned Contriever over its unsupervised counterpart highlights the importance of supervision and task-specific adaptation in dense retrieval.

\subsection{Position Bias: First vs. Last}
\label{sec:position_bias_first_last}
Further evidence is provided in Figure~\ref{fig:initial_final_evidence_ttest}, where we compared two document variants:
\textbf{1.~Beginning-Evidence Document $\bm{D_1}$:} The evidence sentence is positioned at the start of the document.
\textbf{2. End-Evidence Document $\bm{D_2}$:} The same evidence sentence is positioned at the end of the document.

\begin{equation}
\small
\begin{aligned}
    \bm{D_1}& \coloneq \bm{S_{ev}}+\sum_{S^{-h}_{-t} \in D_{orig}} S^{-h}_{-t} \\
     \bm{D_2}& \coloneq \sum_{S^{-h}_{-t} \in D_{orig}} S^{-h}_{-t} + \bm{S_{ev}}
\end{aligned}
\end{equation}

An example of the document pairs (Position Bias) is shown in Table~\ref{tab:examples}.
The resulting t-statistics (Figure~\ref{fig:radar_plot} and \ref{fig:initial_final_evidence_ttest}), where higher positive values indicate a stronger preference for evidence at the beginning ($D_1$) over the end ($D_2$), provide another clear metric of positional bias. These results serve as a foundation for our subsequent analysis in the interplay between biases section.

\input{tables/query_templates}

\input{figs/literal_ttest}
\input{figs/initial_final_evidence_ttest}
\input{figs/repetition_ttest}
\input{figs/brevity_ttest}

\input{tables/examples_poison}
\input{tables/rag_prompts}
\input{tables/literal_bias_complete}

\input{figs/repetition_bias_heatmap}
\input{figs/repeation_bias_heatmap_support}
\input{tables/poison_acc}

\clearpage

\input{figs/decompx_heatmap_3442}
\input{figs/decompx_heatmap_2270}
\input{figs/decompx_heatmap_864}

\end{document}

%% file: figs/radar_plot.tex
\begin{figure}[t!]
\centering
    \includegraphics[width=0.49\textwidth, trim=30 10 15 10, clip]{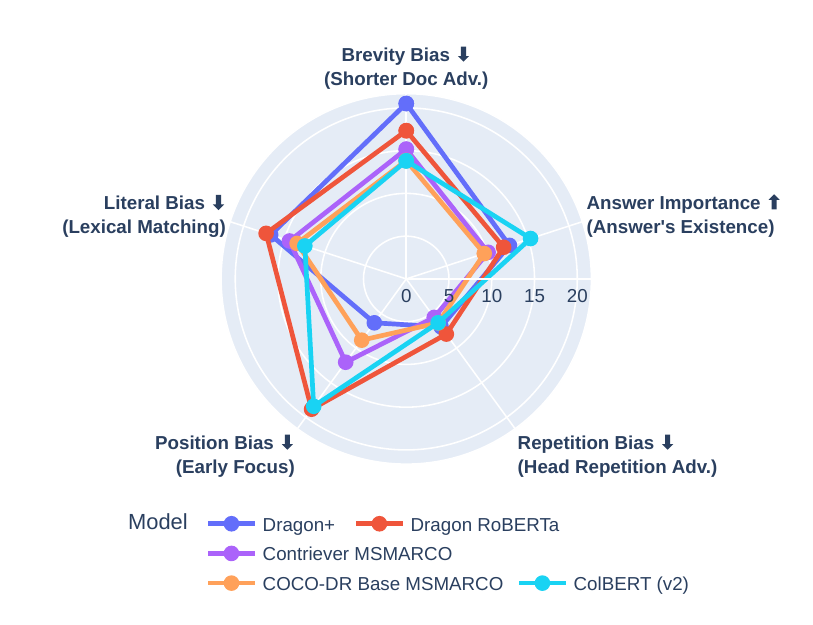}
    \caption{
    Paired t-test statistics comparing retriever scores between document pairs ($D_1$ vs. $D_2$ $\approx M(Query, D_1)-M(Query, D_2)$ where $M$ is the retrieval score of the model). Document pairs are designed for controlled experiments shown in Table~\ref{tab:examples}. Positive values indicate a retriever’s preference for the more biased document in each bias scenario.
    The significance of the answer's existence is often less than the significance of the distracting signals.\protect\footnotemark
    }
    \label{fig:radar_plot}
\end{figure}


%% file: tables/examples.tex
\begin{table*}[t]
\centering
\tiny
\tabcolsep=0.10cm
\begin{tabular}{p{0.08cm}p{0.2cm}p{7.3cm}@{\ \ \ \ }p{7.3cm}}
\toprule 
  && \textbf{Document 1 (Higher Query Document Similarity Score)} - $\bm{D_1}$ & \textbf{Document 2 (Lower Query Document Similarity Score)} - $\bm{D_2}$ \\
  
\midrule 
\addlinespace[0.6em]

\multirow{4}{*}{\rotatebox{90}{\scriptsize{\textbf{Answer}}}}
&\multirow{4}{*}{\rotatebox{90}{\scriptsize{\textbf{Impact}}}}
& \textbf{Query:} What is the sister city of \head{Leonessa}? 
\newline \textbf{Document:} \head{Leonessa} \evidence{is twinned with the French town of} \tail{Gonesse} \evidence{.} \newline Its population in 2008 was around 2,700 . Situated in a small plain at the foot of .....
& \textbf{Query:} What is the sister city of \head{Leonessa}?
\newline \textbf{Document:} \head{Leonessa} is a town and comune in the far northeastern part of the Province of Rieti in the Lazio region of central Italy . \newline Its population in 2008 was around 2,700 . Situated in a small plain at the foot of .....
\\
\addlinespace[0.1em]
\midrule[0.03em]
\addlinespace[0.6em]

\multirow{4}{*}{\rotatebox{90}{\scriptsize{\textbf{Position}}}}
&\multirow{4}{*}{\rotatebox{90}{\scriptsize{\textbf{Bias}}}}
& \textbf{Query:} Which country is \head{Wonyong Sung} a citizen of?
\newline \textbf{Document:} \head{Wonyong Sung} \evidence{( born 1950s ) ,} \tail{South Korean} \evidence{professor of electronic engineering} Won - yong is a Korean masculine given name ..... People with this name include : Kang Won - yong ( 1917 – 2006 ) ..... , South Korean swimmer& \textbf{Query:} Which country is \head{Wonyong Sung} a citizen of?
\newline \textbf{Document:} Won - yong is a Korean masculine given name ..... People with this name include : .... Jung Won - yong ( born 1992 ) , South Korean swimmer \head{Wonyong Sung} \evidence{( born 1950s ) ,} \tail{South Korean} \evidence{professor of electronic engineering}
\\

\addlinespace[0.3em]
\midrule[0.03em]
\addlinespace[0.6em]
\multirow{3}{*}{\rotatebox{90}{\scriptsize{\textbf{Literal}}}}
&\multirow{3}{*}{\rotatebox{90}{\scriptsize{\textbf{Bias}}}}
& \textbf{Query:} When was \head{Seyhun} born?
\newline \textbf{Document:} \head{Seyhun} \evidence{, ( } \tail{August 22 , 1920} \evidence{– May 26 , 2014 ) was an Iranian architect , sculptor , painter , scholar and professor .} He studied fine arts at .....
& \textbf{Query:} When was \head{Seyhun} born?
\newline \textbf{Document:} \head{Houshang Seyhoun} \evidence{, ( } \tail{August 22 , 1920} \evidence{– May 26 , 2014 ) was an Iranian architect , sculptor , painter , scholar and professor .} He studied fine arts at .....
\\

\addlinespace[0.3em]
\midrule[0.03em]
\addlinespace[0.6em]
\multirow{5}{*}{\rotatebox{90}{\scriptsize{\textbf{Brevity}}}}
&\multirow{5}{*}{\rotatebox{90}{\scriptsize{\textbf{Bias}}}}
& \textbf{Query:} What series is \head{Lost Verizon} part of?
\newline \textbf{Document:} \evidence{"} \head{Lost Verizon} \evidence{" is the second episode of} \tail{The Simpsons} \evidence{' twentieth season .}
& \textbf{Query:} What series is \head{Lost Verizon} part of?
\newline \textbf{Document:} \evidence{"} \head{Lost Verizon} \evidence{" is the second episode of} \tail{The Simpsons} \evidence{' twentieth season .} It first aired on the Fox network in the United States on October 5 , 2008 . Bart becomes jealous of his friends and their cell phones . Working at a golf course , Bart takes the cell phone of Denis Leary ..... 
\\
\addlinespace[0.3em]
\midrule[0.03em]
\addlinespace[0.6em]
\multirow{5}{*}{\rotatebox{90}{\scriptsize{\textbf{Repetition}}}}
&\multirow{5}{*}{\rotatebox{90}{\scriptsize{\textbf{Bias}}}}
& \textbf{Query:} Where was \head{James Paul Maher} born?
\newline \textbf{Document:} \evidence{Born in} \tail{Brooklyn , New York} \evidence{,} \head{Maher} \evidence{graduated from St. Patrick 's Academy in Brooklyn .} \head{James Paul Maher} ( November 3 , 1865 – July 31 , 1946 ) was a U.S. Representative from New York . \head{Maher} was elected as a Democrat to the Sixty - second and to the four succeeding Congresses ( March 4 , 1911 – March 4 , 1921 ) .	
& \textbf{Query:} Where was \head{James Paul Maher} born?
\newline \textbf{Document:} \evidence{Born in} \tail{Brooklyn , New York} \evidence{,} \head{Maher} \evidence{graduated from St. Patrick 's Academy in Brooklyn .} Apprenticed to the hatter 's trade , he moved to Danbury , Connecticut in 1887 and was employed as a journeyman hatter . He became treasurer of the United Hatters of North America in 1897 .
\\
\addlinespace[0.3em]
\midrule[0.03em]
\addlinespace[0.6em]
\multirow{5}{*}{\rotatebox{90}{\scriptsize{\textbf{Foil vs. }}}}
&\multirow{5}{*}{\rotatebox{90}{\scriptsize{\textbf{Evidence}}}}
& \textbf{Query:} Who is the publisher of \head{Assassin 's Creed Unity}? 
\newline \textbf{Document:} " \head{Assassin 's Creed Unity} " " \head{Assassin 's Creed Unity} " \head{Assassin 's Creed Unity} received mixed reviews upon its release .
& \textbf{Query:} Who is the publisher of \head{Assassin 's Creed Unity}? 
\newline \textbf{Document:} Isa is a town and Local Government Area in the state of Sokoto in Nigeria . It shares borders with ..... \head{Assassin 's Creed Unity} \evidence{is an action - adventure video game developed by Ubisoft Montreal and published by} \tail{Ubisoft}. Isa is a town and Local Government Area in the state of Sokoto in Nigeria . It shares borders with .....
\\
\addlinespace[0.1em]
\bottomrule
\end{tabular}

\caption{
Construction of query–document pairs. From each Re-DocRED article, we select an \evidence{Evidence Sentence ($S_{ev}$)} that contains both the \head{Head Entity ($h$, the relation’s subject)} and \tail{Tail Entity ($t$, its object)}. We then create two variants: $D_1$, engineered to amplify one bias, and $D_2$, its control. (For position bias, e.g., $D_1$ is ranked higher than $D_2$ because $s_{ev}$ is placed at the top of $D_1$ but at the end of $D_2$.) Paired score differences $M(Q,D_1)-M(Q,D_2)$ over 250 queries provide the paired t-statistics plotted in Fig~\ref{fig:radar_plot}. Only in the Answer-Impact $D_2$ and Foil-vs.-Evidence $D_1$, $t$ is removed, making them non-relevant; in all other tests, both documents contain the answer and remain relevant. In all cases, retrievers favor $D_1$ over $D_2$.
}
\label{tab:examples}
\end{table*}

%% file: figs/decompx_heatmap.tex
\begin{figure*}[t!]
\centering
    \includegraphics[width=0.96\textwidth, trim=0 0 0 0, clip]{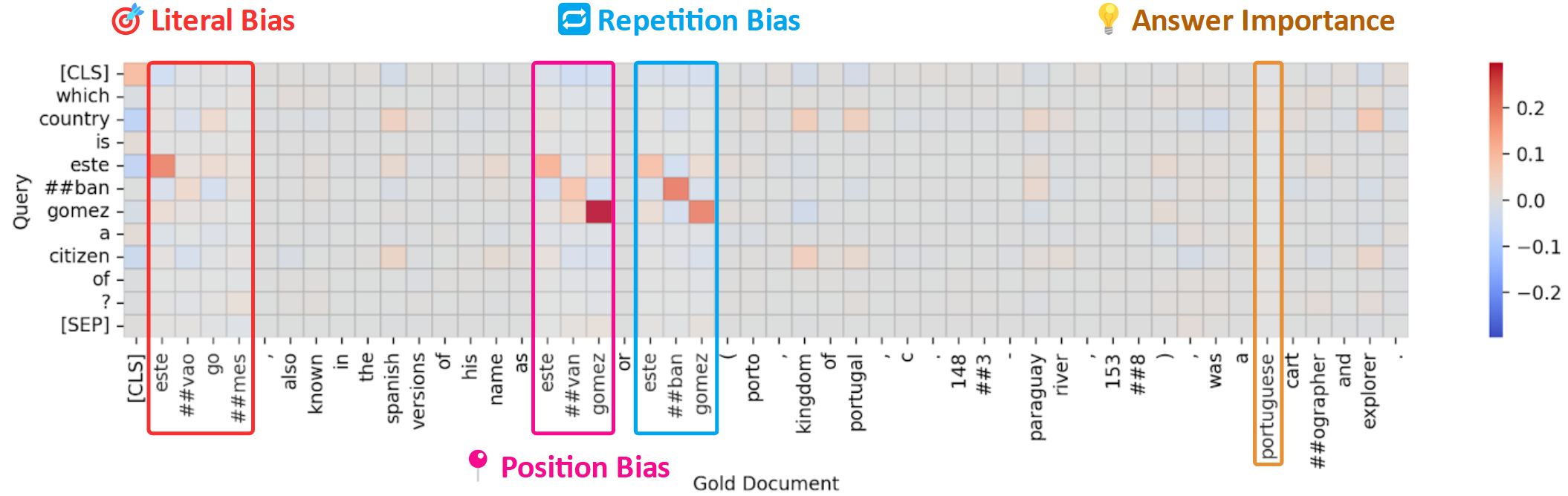}
    \caption{
    Visualization of the contribution of each query and document token to the final retrieval score using DecompX. Literal Bias reflects the model's preference for exact word matches, such as failing to match "esteban goemz" with "estevao gomes." Position Bias indicates a preference for entities earlier in the document receiving more attention. Repetition Bias shows that repeating an entity multiple times increases its score. Lastly, Answer Importance demonstrates that the query's answer entity receives less attention compared to head entity matches.
    }
    \label{fig:decompx_heatmap}
\end{figure*}


%% file: tables/models_performance_nq.tex
\begin{table}[t]
\centering
\scriptsize
\tabcolsep=0.15cm

\begin{tabular}{llrr}
\toprule
Model & pooling & nDCG@10 & Recall@10 \\
\midrule
Dragon RoBERTa  & cls & \textbf{0.55} & \textbf{0.75} \\
Dragon+  & cls & 0.54 & 0.74 \\
COCO-DR Base MSMARCO & cls & 0.50 & 0.71 \\
Contriever MSMARCO & avg & 0.50 & 0.71 \\
RetroMAE MSMARCO FT & cls & 0.48 & 0.68 \\
Contriever  & avg & 0.25 & 0.41 \\
\bottomrule
\end{tabular}

\caption{
Models' performance on NQ dataset with test set queries and 2,681,468 corpus size.
} 
\label{tab:models_performance_nq}
\end{table}

%% file: figs/tail_ttest.tex
\begin{figure}[t!]
\centering
    \includegraphics[width=0.48\textwidth, trim=0 0 0 0, clip]{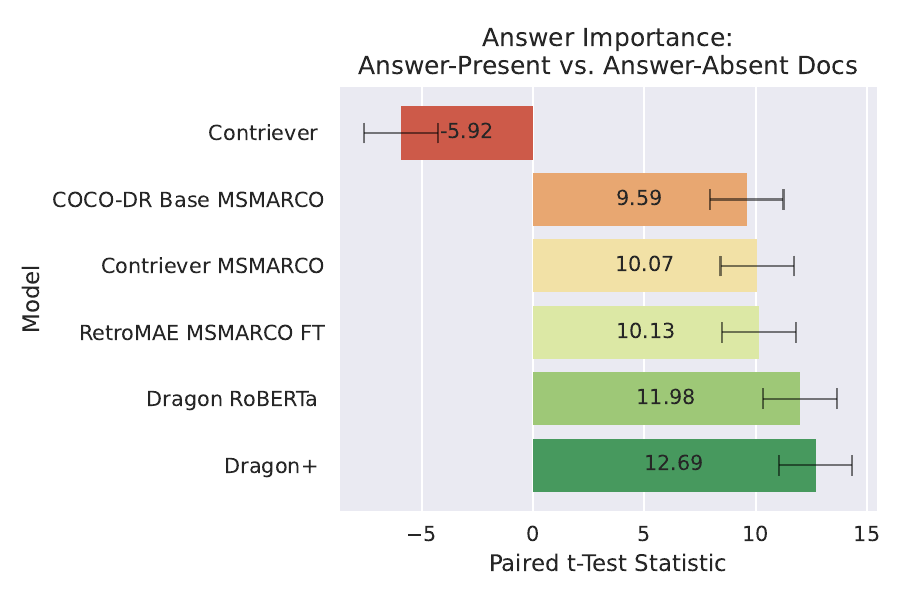}
    \caption{
    Paired t-test statistics comparing dot product similarity between the first sentence containing both head and tail (Answer) entities versus only the head entity, with 95\% CI error bars. Higher values indicate recognition of the answer's importance.
    }
    \label{fig:tail_ttest}
\end{figure}

%% file: figs/position_evidence_dot.tex
\begin{figure}[t!]
\centering
    \includegraphics[width=0.49\textwidth, trim=0 0 0 0, clip]{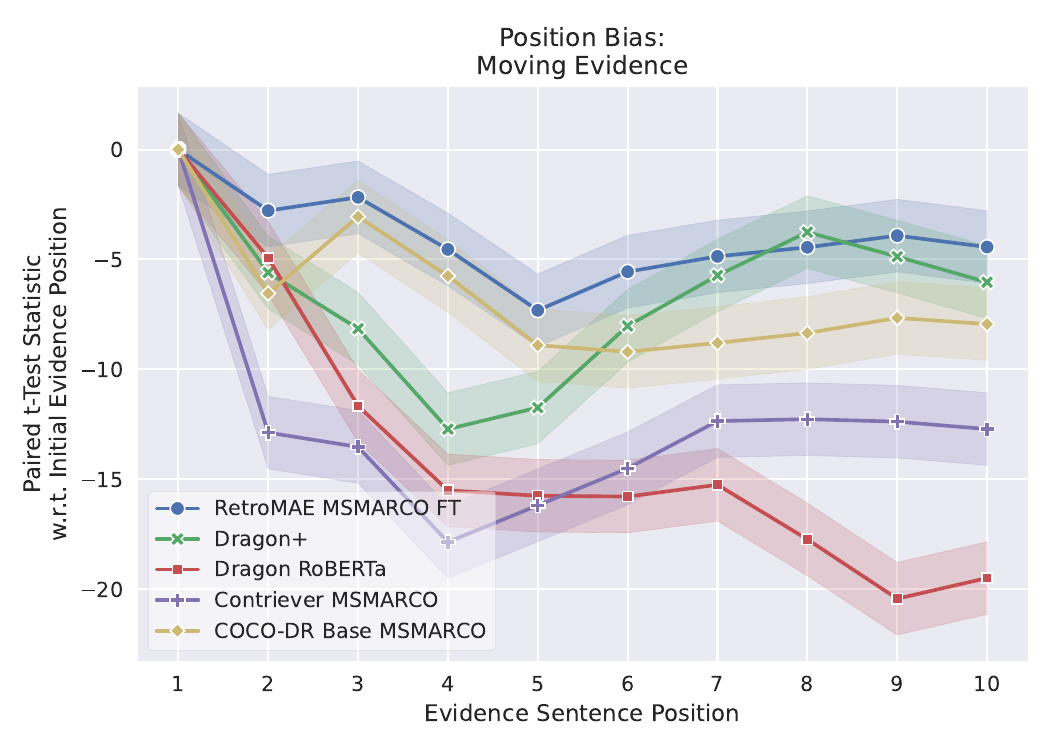}
    \caption{
    Paired t-test statistics comparing the effect of moving the evidence sentence position within the document to keeping it in the first position. Negative values indicate a bias towards the beginning of the document.
    }
    \label{fig:position_evidence_dot}
\end{figure}


%% file: tables/literal_bias.tex
\begin{table}[t]
\centering
\small
\tabcolsep=0.12cm

\begin{tabular}{ccccrr}
\toprule
 &  &  & Model & \makecell{Contriever \\ MSMARCO} & \makecell{Dragon+ \\ } \\
\makecell{$Q1$} & \makecell{$D1$} & \makecell{$Q2$} & \makecell{$D2$} &  &  \\
\midrule
\multirow[t]{3}{*}{\textbf{long}} & \multirow[t]{3}{*}{\textbf{long}} & \textbf{long} & \textbf{short} & +21.05 & +21.04 \\

\textbf{} & \textbf{} & \multirow[t]{2}{*}{\textbf{short}} & \textbf{long} & +22.04 & +13.40 \\

\midrule

\multirow[t]{2}{*}{\textbf{short}} & \multirow[t]{2}{*}{\textbf{short}} & \textbf{long} & \textbf{short} & +4.62 & +9.04 \\

\textbf{} & \textbf{} & \textbf{short} & \textbf{long} & +14.37 & +16.62 \\

\bottomrule
\end{tabular}

\caption{
Paired t-test statistics (p-values < 0.05) comparing retrieval scores between exact name matches ($Q1$-$D1$) and variant name pairs ($Q2$-$D2$). Positive statistics indicate a preference for exact literal matches over semantically equivalent name variants (e.g., ``US''-``US'' over ``US''-``United States''). (All models in Table~\ref{tab:literal_bias_complete}.)
} 
%
\label{tab:literal_bias}
\end{table}

%% file: figs/repeatition_bias_heatmap_contriever.tex
\begin{figure}[t!]
\centering
    \includegraphics[width=0.48\textwidth, trim=0 10 0 8, clip]{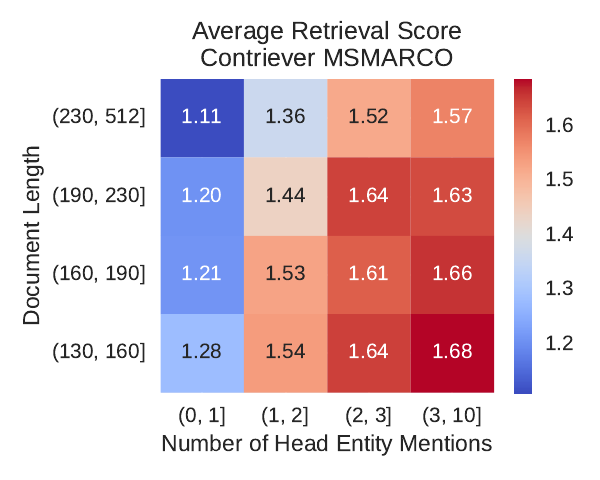}
    \caption{
    The average retrieval score of Contriever MSMARCO increases with head entity repetitions but decreases with document length (all models in Figure~\ref{fig:repetition_bias_heatmap}).
    }
    \label{fig:repetition_bias_heatmap_contriever}
\end{figure}

%% file: tables/foil_acc.tex
\begin{table}[t]
\centering
\small
\tabcolsep=0.1mm

\begin{tabular}{lccc}
\toprule
Model & Accuracy & \makecell{Paired t-Test \\ Statistic} & p-value \\
\midrule
Contriever  & \textcolor{DarkRed}{0.4\%} & -34.58 & < 0.01 \\
RetroMAE MSMARCO FT & \textcolor{DarkRed}{0.4\%} & -41.49 & < 0.01 \\
Contriever MSMARCO & \textcolor{DarkRed}{0.8\%} & -42.25 & < 0.01 \\
Dragon RoBERTa  & \textcolor{DarkRed}{0.8\%} & -36.53 & < 0.01 \\
Dragon+  & \textcolor{DarkRed}{1.2\%} & -40.94 & < 0.01 \\
COCO-DR Base MSMARCO & \textcolor{DarkRed}{2.4\%} & -32.92 & < 0.01 \\
ColBERT (v2) & \textcolor{DarkRed}{7.6\%} & -20.96 & < 0.01 \\
ReasonIR-8B & \textcolor{DarkRed}{8.0\%} & -36.92 & < 0.01 \\
\bottomrule
\end{tabular}

\caption{
The accuracy and paired t-test comparing a foil document $D_1$ (exploiting biases but lacking the answer) to a second document $D_2$ with evidence embedded in unrelated sentences. Accuracy is the proportion of 250 example pairs where $M(Q,D_2) > M(Q,D_1)$. All retrieval models perform extremely poorly (\textcolor{DarkRed}{<10\%} accuracy), highlighting their inability to distinguish biased distractors from genuine evidence. (Extra models explained in \S\ref{sec:more_models})
}
\label{tab:foil_acc}
\end{table}

%% file: tables/rag.tex
\begin{table}[t]
\centering
\tabcolsep=1.3mm


\begin{tabular}{lcccc}
\toprule
\textsc{} & Poison & Foil & No & Evidence \\
\textsc{Model} & Doc & Doc & Doc & Doc \\
\midrule
gpt-4o-mini & \textcolor{DarkRed}{32.0\%} & 44.0\% & 52.0\% & \textbf{88.0\%} \\
gpt-4o & \textcolor{DarkRed}{30.8\%} & 62.8\% & 64.8\% & \textbf{93.6\%} \\
\bottomrule
\end{tabular}

\caption{
RAG accuracy when using different document versions as references. The poisoned document, preferred by retrievers 100\% of the time (Table~\ref{tab:poison_acc}), results in worse performance than providing no document, highlighting the impact of retriever biases on RAG.
}
\label{tab:rag}
\end{table}

%% file: tables/foil_acc_more.tex
\begin{table}[t]
\centering
\small
\tabcolsep=0.1mm

\begin{tabular}{lccc}
\toprule
Model & Accuracy & \makecell{Paired t-Test \\ Statistic} & p-value \\
\midrule
Contriever  & \textcolor{DarkRed}{0.4\%} & -34.58 & < 0.01 \\
RetroMAE MSMARCO FT & \textcolor{DarkRed}{0.4\%} & -41.49 & < 0.01 \\
Contriever MSMARCO & \textcolor{DarkRed}{0.8\%} & -42.25 & < 0.01 \\
Dragon RoBERTa  & \textcolor{DarkRed}{0.8\%} & -36.53 & < 0.01 \\
Dragon+  & \textcolor{DarkRed}{1.2\%} & -40.94 & < 0.01 \\
COCO-DR Base MSMARCO & \textcolor{DarkRed}{2.4\%} & -32.92 & < 0.01 \\
ColBERT (v2) & \textcolor{DarkRed}{7.6\%} & -20.96 & < 0.01 \\
ReasonIR-8B & \textcolor{DarkRed}{8.0\%} & -36.92 & < 0.01 \\
\bottomrule
\end{tabular}

\caption{
The accuracy and paired t-test comparing a foil document (exploiting biases but lacking the answer) to a second document with evidence embedded in unrelated sentences. All retrieval models perform extremely poorly, highlighting their inability to distinguish biased distractors from genuine evidence.
}
\label{tab:foil_acc_more}
\vspace{-1em}
\end{table}

%% file: tables/models.tex
\begin{table}[h]
\scriptsize
\centering
\begin{tabular}{lc}
    \toprule
    \multicolumn{1}{c} {Model} & \multicolumn{1}{c} {Citation} \\
    \midrule
    \href{https://huggingface.co/facebook/dragon-plus-query-encoder}{facebook/dragon-plus-query-encoder} & \citet{lin-etal-2023-train} \\
    \href{https://huggingface.co/facebook/dragon-plus-context-encoder}{facebook/dragon-plus-context-encoder} & \\
    \midrule
    \href{https://huggingface.co/facebook/dragon-roberta-query-encoder}{facebook/dragon-roberta-query-encoder} & \citet{lin-etal-2023-train}\\
    \href{https://huggingface.co/facebook/dragon-roberta-context-encoder}{facebook/dragon-roberta-context-encoder} & \\
    \midrule
    \href{https://huggingface.co/facebook/contriever-msmarco}{facebook/contriever-msmarco} & \citet{izacard2021contriever} \\
    \midrule
    \href{https://huggingface.co/facebook/contriever}{facebook/contriever} & \citet{izacard2021contriever}\\
    \midrule
    \href{https://huggingface.co/OpenMatch/cocodr-base-msmarco}{OpenMatch/cocodr-base-msmarco} & \citet{yu-etal-2022-coco}\\
    \midrule
    \href{https://huggingface.co/Shitao/RetroMAE_MSMARCO_finetune}{Shitao/RetroMAE\_MSMARCO\_finetune} & \citet{xiao-etal-2022-retromae}\\
    \midrule

    \href{https://huggingface.co/colbert-ir/colbertv2.0}{colbert-ir/colbertv2.0} & \citet{10.1145/3397271.3401075-colbert} \\

    \midrule
    \href{https://huggingface.co/reasonir/ReasonIR-8B}{reasonir/ReasonIR-8B} & \citet{shao2025reasonirtrainingretrieversreasoning} \\
    \midrule
    
    gpt-4o-mini-2024-07-18 & \citet{openai2024gpt4ocard} \\
    \midrule
    gpt-4o-2024-08-06 & \citet{openai2024gpt4ocard} \\
    \bottomrule
\end{tabular}
\caption{The details of the models we used in this work.}
\label{tab:models}
\end{table}

%% file: tables/models_performance.tex
\begin{table}[h]
\centering
\scriptsize
\tabcolsep=0.15cm

\begin{tabular}{llrr}
\toprule
Model & Pooling & nDCG@10 & Recall@10 \\
\midrule
Dragon+  & cls & \textbf{0.55} & \textbf{0.63} \\
Dragon RoBERTa  & cls & 0.53 & 0.59 \\
Contriever MSMARCO & avg & 0.52 & 0.59 \\
Contriever  & avg & 0.50 & 0.59 \\
RetroMAE MSMARCO FT & cls & 0.49 & 0.55 \\
COCO-DR Base MSMARCO & cls & 0.48 & 0.53 \\
\bottomrule
\end{tabular}

\caption{
Models' performance on our refined redocred dataset with 7170 queries and 105925 corpus size.
} 
\label{tab:models_performance}
\end{table}

%% file: tables/preliminary_analysis.tex
\begin{table}[h]
\centering
\scriptsize
\tabcolsep=0.10cm

\begin{tabular}{lrr}
\toprule
Issue & Count & Percentage \\
\midrule
Long Document & 33 & 55\% \\
Missing Answer & 19 & 32\% \\
Literal Bias & 11 & 18\% \\
Repetition & 6 & 10\% \\
Numbers & 2 & 3\% \\
Position Bias & 2 & 3\% \\
\bottomrule
\end{tabular}

\caption{
Preliminary findings from our annotation of 60 retrieval errors based on DecompX
}
\label{tab:preliminary_analysis}
\end{table}

%% file: tables/query_templates.tex
\begin{table*}[h]
\centering
\scriptsize
\begin{tabular}{lll}
\toprule 
\textbf{Relation ID} & \textbf{Relation Name} & \textbf{Query Template} \\
\midrule 

P131 & located in the administrative territorial entity & Which administrative territorial entity is <head\_entity> located in? \\\midrule
P577 & publication date & When was <head\_entity> published? \\\midrule
P17 & country & Which country is <head\_entity> associated with? \\\midrule
P264 & record label & Which record label is <head\_entity> associated with? \\\midrule
P571 & inception & When was <head\_entity> founded? \\\midrule
P361 & part of & What is <head\_entity> a part of? \\\midrule
P800 & notable work & What is a notable work of <head\_entity>? \\\midrule
P569 & date of birth & When was <head\_entity> born? \\\midrule
P159 & headquarters location & Where is the headquarters of <head\_entity> located? \\\midrule
P527 & has part & What are the components of <head\_entity>? \\\midrule
P123 & publisher & Who is the publisher of <head\_entity>? \\\midrule
P175 & performer & Who performed <head\_entity>? \\\midrule
P449 & original network & What is the original network of <head\_entity>? \\\midrule
P706 & located on terrain feature & Where is <head\_entity> located on a terrain feature? \\\midrule
P580 & start time & When did <head\_entity> start? \\\midrule
P740 & location of formation & Where was <head\_entity> formed? \\\midrule
P27 & country of citizenship & Which country is <head\_entity> a citizen of? \\\midrule
P403 & mouth of the watercourse & What is the mouth of the watercourse of <head\_entity>? \\\midrule
P570 & date of death & When did <head\_entity> die? \\\midrule
P136 & genre & What genre does <head\_entity> belong to? \\\midrule
P576 & dissolved, abolished or demolished & When was <head\_entity> dissolved or demolished? \\\midrule
P495 & country of origin & What is the country of origin of <head\_entity>? \\\midrule
P19 & place of birth & Where was <head\_entity> born? \\\midrule
P155 & follows & What precedes <head\_entity>? \\\midrule
P400 & platform & What platform is <head\_entity> available on? \\\midrule
P1344 & participant of & What was <head\_entity> a participant of? \\\midrule
P3373 & sibling & Who is the sibling of <head\_entity>? \\\midrule
P676 & lyrics by & Who wrote the lyrics for <head\_entity>? \\\midrule
P26 & spouse & Who is the spouse of <head\_entity>? \\\midrule
P58 & screenwriter & Who wrote the screenplay for <head\_entity>? \\\midrule
P35 & head of state & Who is the head of state of <head\_entity>? \\\midrule
P6 & head of government & Who is the head of government of <head\_entity>? \\\midrule
P178 & developer & Who developed <head\_entity>? \\\midrule
P279 & subclass of & What is <head\_entity> a subclass of? \\\midrule
P127 & owned by & Who owns <head\_entity>? \\\midrule
P156 & followed by & What follows <head\_entity>? \\\midrule
P140 & religion & What is the religion of <head\_entity>? \\\midrule
P607 & conflict & What conflict was <head\_entity> part of? \\\midrule
P364 & original language of work & What is the original language of <head\_entity>? \\\midrule
P463 & member of & Which organization is <head\_entity> a member of? \\\midrule
P179 & series & What series is <head\_entity> part of? \\\midrule
P176 & manufacturer & Who manufactured <head\_entity>? \\\midrule
P190 & sister city & What is the sister city of <head\_entity>? \\\midrule
P20 & place of death & Where did <head\_entity> die? \\\midrule
P112 & founded by & Who founded <head\_entity>? \\\midrule
P31 & instance of & What is <head\_entity> an instance of? \\\midrule
P276 & location & Where is <head\_entity> located? \\\midrule
P86 & composer & Who composed the music for <head\_entity>? \\\midrule
P57 & director & Who directed <head\_entity>? \\\midrule
P272 & production company & Which production company produced <head\_entity>? \\\midrule
P50 & author & Who is the author of <head\_entity>? \\

\bottomrule
\end{tabular}
\caption{
The query templates for each relation.
} 
\label{tab:query_templates}
\end{table*}

%% file: figs/literal_ttest.tex
\begin{figure}[h]
\centering
    \includegraphics[width=0.49\textwidth, trim=0 0 0 0, clip]{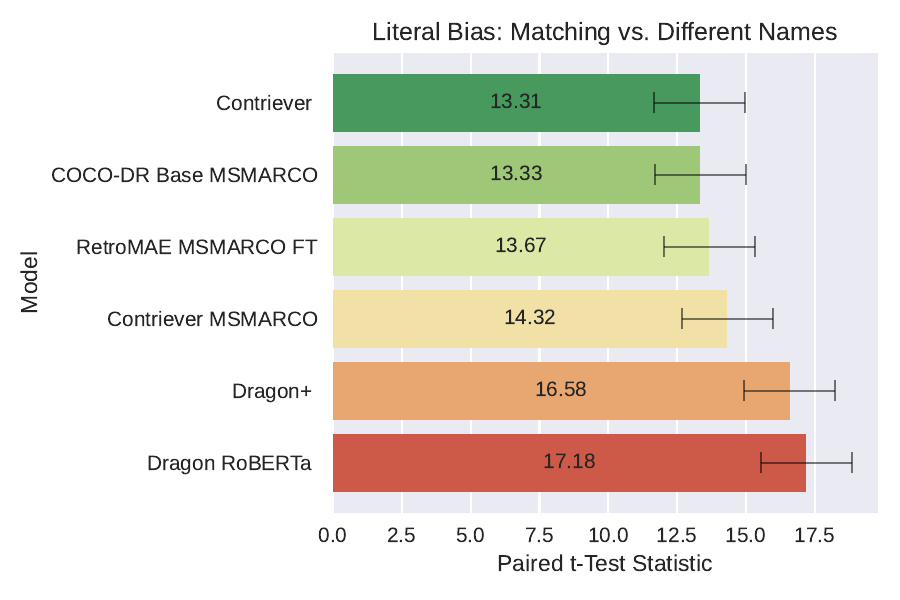}
    \caption{
    Paired t-test statistics comparing retrieval scores between two scenarios: (1) when both query and document use the shortest name variant, and (2) when the query uses the short name but the document contains the long name variant of the same entity. Positive statistics indicate that models favor exact string matches over semantic matching of equivalent entity names.
    }
    \vspace{-1em}
    \label{fig:literal_ttest}
\end{figure}

%% file: figs/initial_final_evidence_ttest.tex
\begin{figure}[h]
\centering
    \includegraphics[width=0.49\textwidth, trim=0 0 0 0, clip]{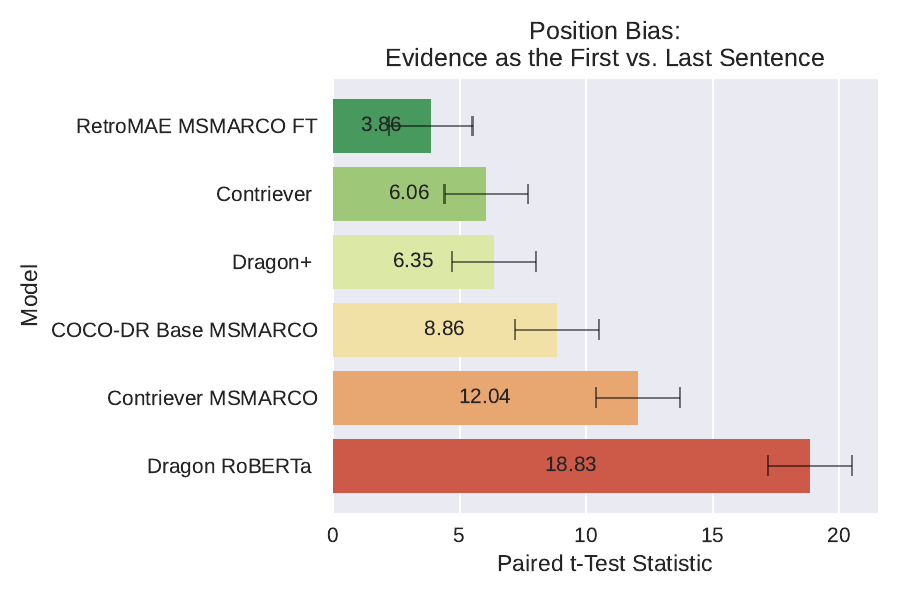}
    \caption{
    Paired t-test statistics comparing document scores based on the position of the evidence sentence (beginning vs. end). Higher positive values reflect a preference for evidence at the beginning, indicating positional bias.
    }
    \label{fig:initial_final_evidence_ttest}
\end{figure}

%% file: figs/repetition_ttest.tex
\begin{figure}[h]
\centering
    \includegraphics[width=0.49\textwidth, trim=0 0 0 0, clip]{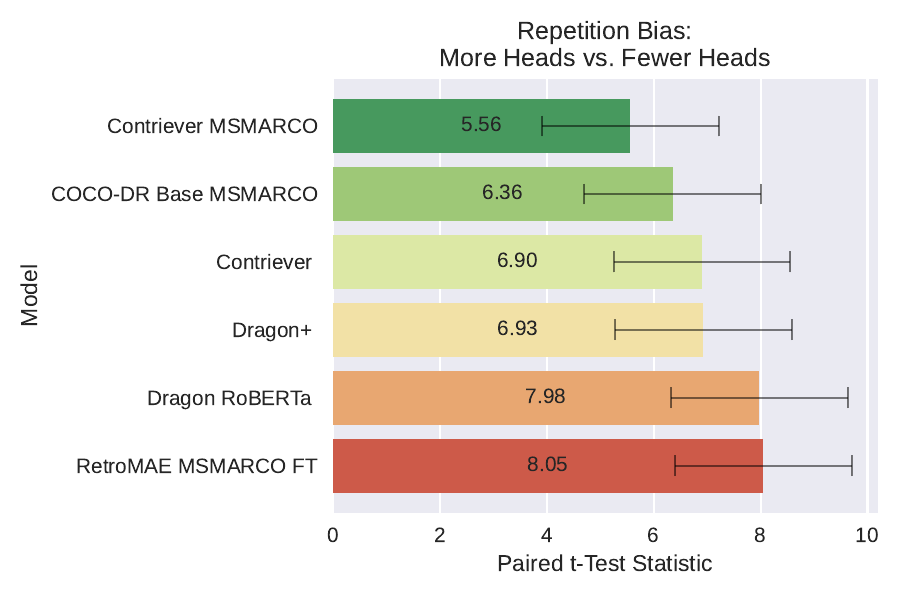}
    \caption{
    Paired t-test statistics comparing the dot product similarity of queries with two sets of sentences: (1) \textit{More Heads}, consisting of evidence and two sentences with head mentions but no tails, and (2) \textit{Fewer Heads}, consisting of evidence and two sentences without head or tail mentions. Positive values indicate higher similarity for sentences with more heads.
    }
    \label{fig:repetition_ttest}
\end{figure}

%% file: figs/brevity_ttest.tex
\begin{figure}[h]
\centering
    \includegraphics[width=0.49\textwidth, trim=0 0 0 0, clip]{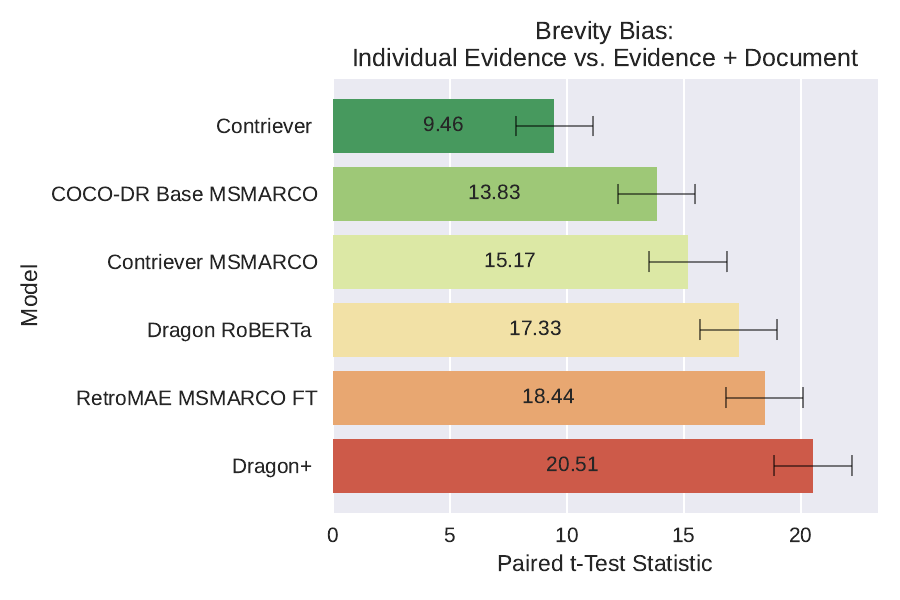}
    \caption{
    Paired t-test statistics comparing scores for documents containing only the evidence sentence versus those containing the evidence plus the full document. Higher positive values indicate a stronger model bias toward brevity.
    }
    \label{fig:brevity_ttest}
\end{figure}

%% file: tables/examples_poison.tex
\begin{table*}[h]
\centering
\tiny
\tabcolsep=0.10cm
\begin{tabular}{lp{7.3cm}@{\ \ \ \ }p{7.3cm}}
\toprule 
  & \textbf{Method 1 (Higher Query Document Similarity Score)} & \textbf{Method 2 (Lower Query Document Similarity Score)} \\
\midrule 

\multirow{5}{*}{\rotatebox{90}{Foil vs. Evide.}}
& \textbf{Query:} Who is the publisher of \head{Assassin 's Creed Unity}? 
\newline \textbf{Document:} " \head{Assassin 's Creed Unity} " " \head{Assassin 's Creed Unity} " \head{Assassin 's Creed Unity} received mixed reviews upon its release .
& \textbf{Query:} Who is the publisher of \head{Assassin 's Creed Unity}? 
\newline \textbf{Document:} Isa is a town and Local Government Area in the state of Sokoto in Nigeria . It shares borders with ..... \head{Assassin 's Creed Unity} \evidence{is an action - adventure video game developed by Ubisoft Montreal and published by} \tail{Ubisoft}. Isa is a town and Local Government Area in the state of Sokoto in Nigeria . It shares borders with .....
\\\midrule[0.03em]
\multirow{6}{*}{\rotatebox{90}{Poison vs. Evide.}}
& \textbf{Query:} Who is the publisher of \head{Assassin 's Creed Unity}? 
\newline \textbf{Document:} " \head{Assassin 's Creed Unity} " " \head{Assassin 's Creed Unity} " \head{Assassin 's Creed Unity} received mixed reviews upon its release . \head{Assassin 's Creed Unity} \evidence{is an action - adventure video game developed by Electronic Arts Montreal and published by} \poison{Electronic Arts}
& \textbf{Query:} Who is the publisher of \head{Assassin 's Creed Unity}? 
\newline \textbf{Document:} Isa is a town and Local Government Area in the state of Sokoto in Nigeria . It shares borders with ..... \head{Assassin 's Creed Unity} \evidence{is an action - adventure video game developed by Ubisoft Montreal and published by} \tail{Ubisoft}. Isa is a town and Local Government Area in the state of Sokoto in Nigeria . It shares borders with .....\\\\
\bottomrule
\end{tabular}

\caption{
Examples from our framework for poison document and evidence document highlighting \evidence{Evidence}, \head{Head Entity}, \tail{Tail Entity} and \poison{Poison} replacing true tail entity. In all cases, retrieval models favor Method 1 over Method 2, assigning higher retrieval scores accordingly.
}
\label{tab:examples_poison}
\end{table*}

%% file: tables/rag_prompts.tex
\begin{table*}[h]
\centering
\scriptsize
\begin{tabular}{lp{8.0cm}}
\toprule 
\textbf{Prompt Utility} & \textbf{Prompt} \\
\midrule 

Poisoning & 
In the sentence: '\{evidence\}', replace the entity '\{tail\}' with a different entity that makes sense in context but is completely different. Output only the replacement entity. replacement entity: 
\\
\midrule
RAG & 
Answer the question based on the given document. Only give me the complete answer and do not output any other words. The following is the given document.

Document: \{doc\}

Question: \{query\} 

Answer: 
\\
\midrule
RAG for No Doc & 
Answer the question. Only give me the answer and do not output any other words.

Question: \{query\} 

Answer:
\\
\midrule
Evaluation & 
Query: \{query\}

Evidence: \{evidence\_sentence\}

Gold Answer: \{gold\_answer\}

Model Answer: \{model\_answer\}

Does the Model Answer contain or imply the Gold Answer based on the evidence? YES or NO : 
\\

\bottomrule
\end{tabular}
\caption{
The prompts utilized for RAG.
} 
\label{tab:rag_prompts}
\end{table*}

%% file: tables/literal_bias_complete.tex
\begin{table*}[h]
\centering
\tiny
\tabcolsep=0.12cm

\begin{tabular}{p{1.2cm}p{1.2cm}p{1.2cm}p{1.2cm}rrrrrr}
\toprule
 &  &  & Model & \makecell{COCO-DR \\ Base MSMARCO} & \makecell{RetroMAE \\ MSMARCO FT} & \makecell{Contriever \\ } & \makecell{Contriever \\ MSMARCO} & \makecell{Dragon+ \\ } & \makecell{Dragon \\ RoBERTa} \\
Query Name 1 & Doc Name 1 & Query Name 2 & Doc Name 2 &  &  &  &  &  &  \\
\midrule
\multirow[t]{3}{*}{\textbf{long}} & \multirow[t]{3}{*}{\textbf{long}} & \textbf{long} & \textbf{short} & 20.67 & 21.92 & 19.22 & 21.05 & 21.03 & 21.64 \\
\textbf{} & \textbf{} & \multirow[t]{2}{*}{\textbf{short}} & \textbf{long} & 23.41 & 23.53 & 21.46 & 22.01 & 13.40 & 7.55 \\
\textbf{} & \textbf{} & \textbf{short} & \textbf{short} & 18.43 & 19.60 & 16.41 & 17.35 & 4.99 & 1.75 \\
\midrule
\multirow[t]{2}{*}{\textbf{short}} & \multirow[t]{2}{*}{\textbf{short}} & \textbf{long} & \textbf{short} & 2.19 & 3.86 & 2.32 & 4.65 & 9.05 & 5.57 \\
\textbf{} & \textbf{} & \textbf{short} & \textbf{long} & 13.33 & 13.67 & 13.31 & 14.32 & 16.58 & 17.18 \\
\bottomrule
\end{tabular}

\caption{
Paired t-test statistics comparing retrieval scores between exact name matches (Q1-D1) and variant name pairs (Q2-D2). Positive statistics indicate model preference for exact literal matches over semantically equivalent name variants (e.g., preferring ``US''-``US'' over ``US''-``United States'').
} 
%
\label{tab:literal_bias_complete}
\end{table*}

%% file: figs/repetition_bias_heatmap.tex
\begin{figure*}[h]
\centering
    \includegraphics[width=0.99\textwidth, trim=0 210 0 0, clip]{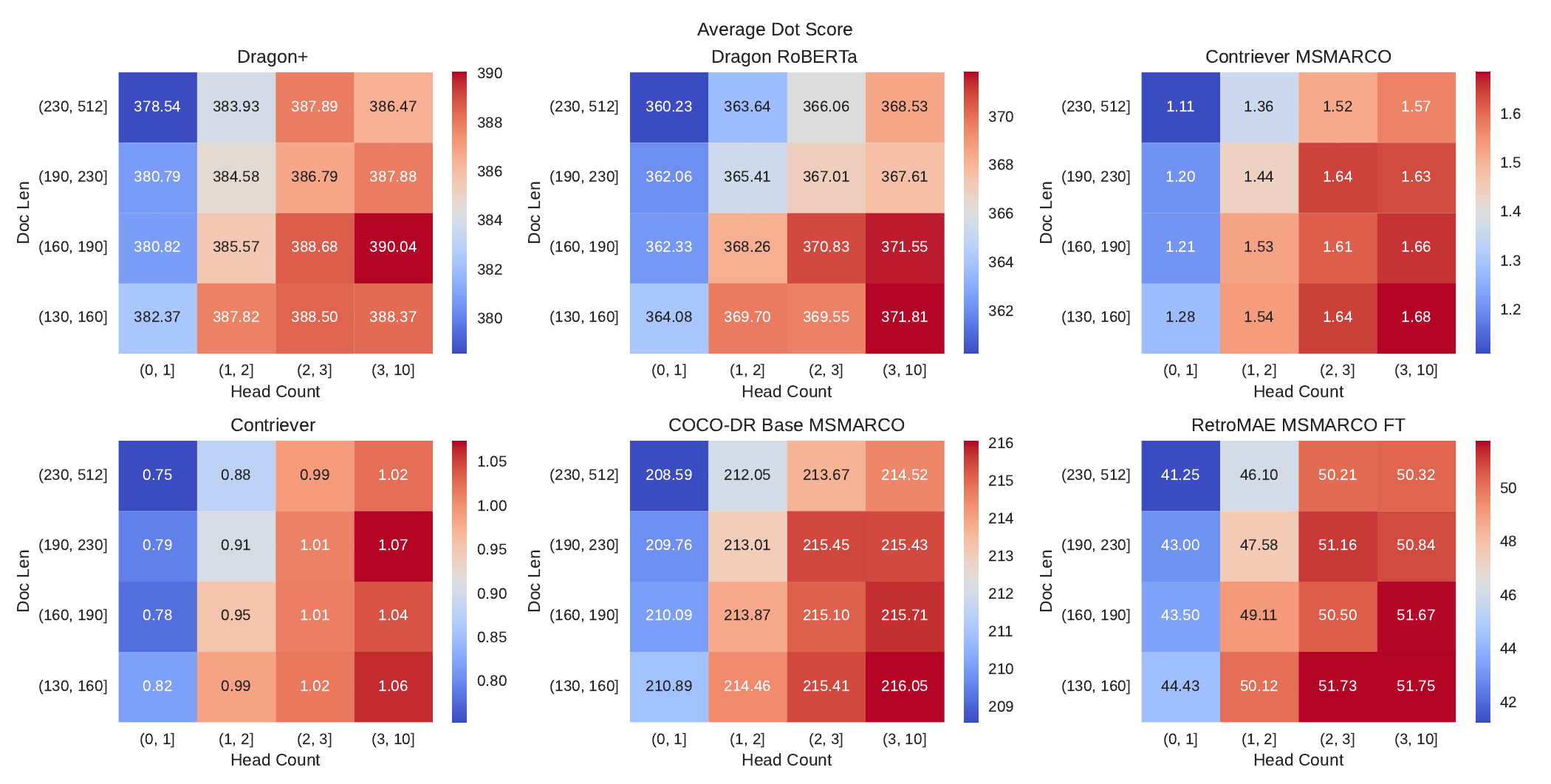}
    \caption{
    The average retrieval dot product score for samples with different document lengths and head entity repetitions. (See Figure~\ref{fig:repetition_bias_heatmap_support} for the number of examples in each bin)
    }
    \vspace{-1em}
    \label{fig:repetition_bias_heatmap}
\end{figure*}

%% file: figs/repeation_bias_heatmap_support.tex
\begin{figure*}[h]
\centering
    \includegraphics[width=0.40\textwidth, trim=0 10 0 0, clip]{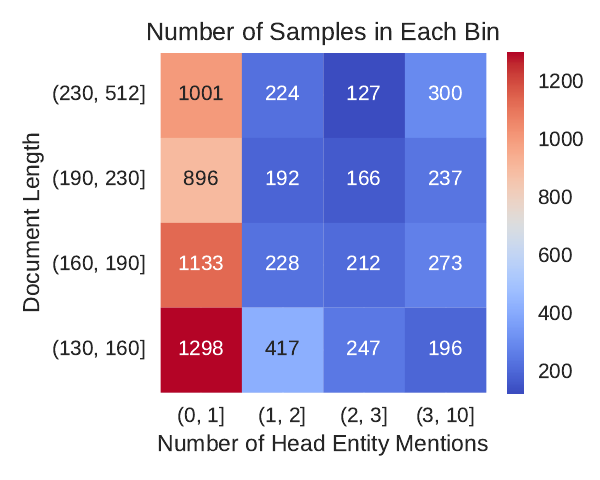}
    \caption{
    The number of samples in each bin of Figures \ref{fig:repetition_bias_heatmap} and \ref{fig:repetition_bias_heatmap_contriever}. 
    }
    \vspace{-1em}
    \label{fig:repetition_bias_heatmap_support}
\end{figure*}

%% file: tables/poison_acc.tex
\begin{table*}[h]
\centering
\scriptsize
\tabcolsep=0.10cm

\begin{tabular}{lrrr}
\toprule
Model & Accuracy & \makecell{Paired t-Test \\ Statistic} & p-value \\
\midrule
Dragon+  & \textcolor{DarkRed}{0.0\%} & -55.16 & < 0.01 \\
Dragon RoBERTa  & \textcolor{DarkRed}{0.0\%} & -49.17 & < 0.01 \\
Contriever MSMARCO & \textcolor{DarkRed}{0.0\%} & -46.96 & < 0.01 \\
COCO-DR Base MSMARCO & \textcolor{DarkRed}{0.0\%} & -40.19 & < 0.01 \\
RetroMAE MSMARCO FT & \textcolor{DarkRed}{0.0\%} & -48.10 & < 0.01 \\
Contriever  & \textcolor{DarkRed}{1.2\%} & -33.60 & < 0.01 \\
\bottomrule
\end{tabular}

\caption{
The accuracy, paired t-test statistics, and p-values comparing a \textbf{poison document}, designed to exploit biases and having a wrong answer (tail), against a second document containing the evidence sentence embedded in the middle of eight unrelated sentences from a different document. All retrieval models perform extremely poorly (less than \textcolor{DarkRed}{2\%} accuracy).
} 
\label{tab:poison_acc}
\end{table*}

%% file: figs/decompx_heatmap_3442.tex
\begin{figure*}[t!]
\centering
    \includegraphics[width=0.99\textwidth, trim=0 0 90 0, clip]{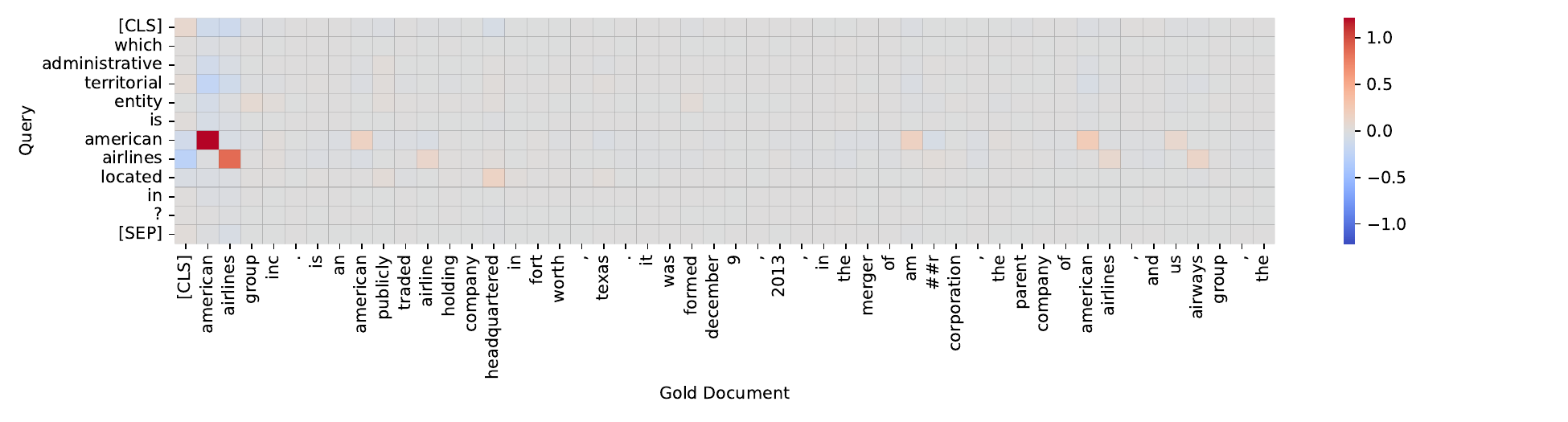}
    \caption{
    Visualization of token-wise effects on retriever scores using DecompX.
    }
    \label{fig:decompx_heatmap_3442}
\end{figure*}

%% file: figs/decompx_heatmap_2270.tex
\begin{figure*}[t!]
\centering
    \includegraphics[width=0.99\textwidth, trim=0 0 90 0, clip]{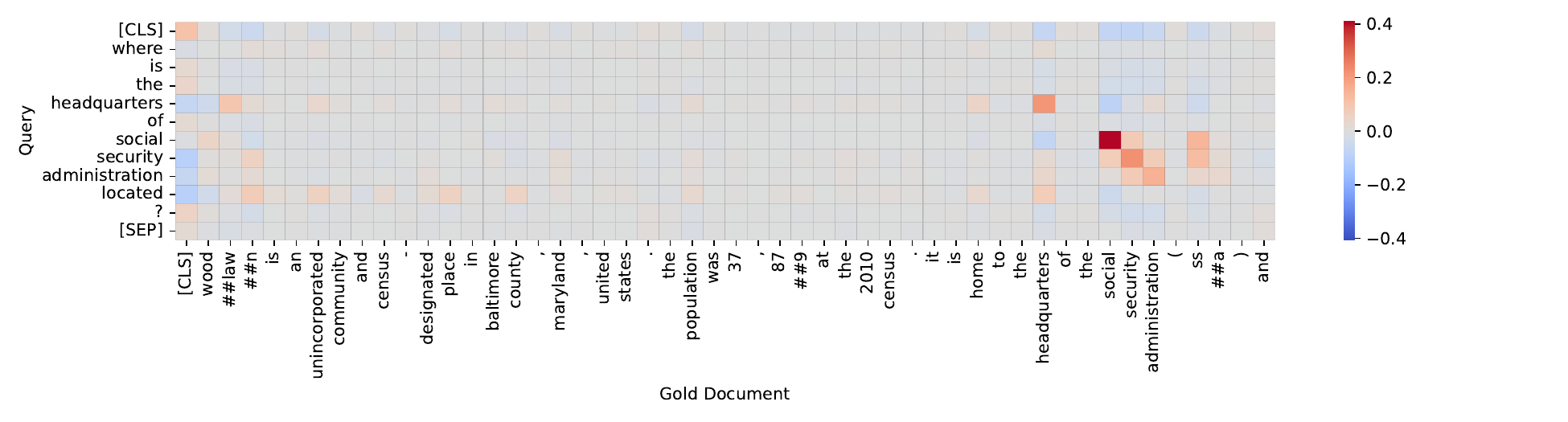}
    \caption{
    Visualization of token-wise effects on retriever scores using DecompX.
    }
    \label{fig:decompx_heatmap_2270}
\end{figure*}

%% file: figs/decompx_heatmap_864.tex
\begin{figure*}[t!]
\centering
    \includegraphics[width=0.99\textwidth, trim=0 0 90 0, clip]{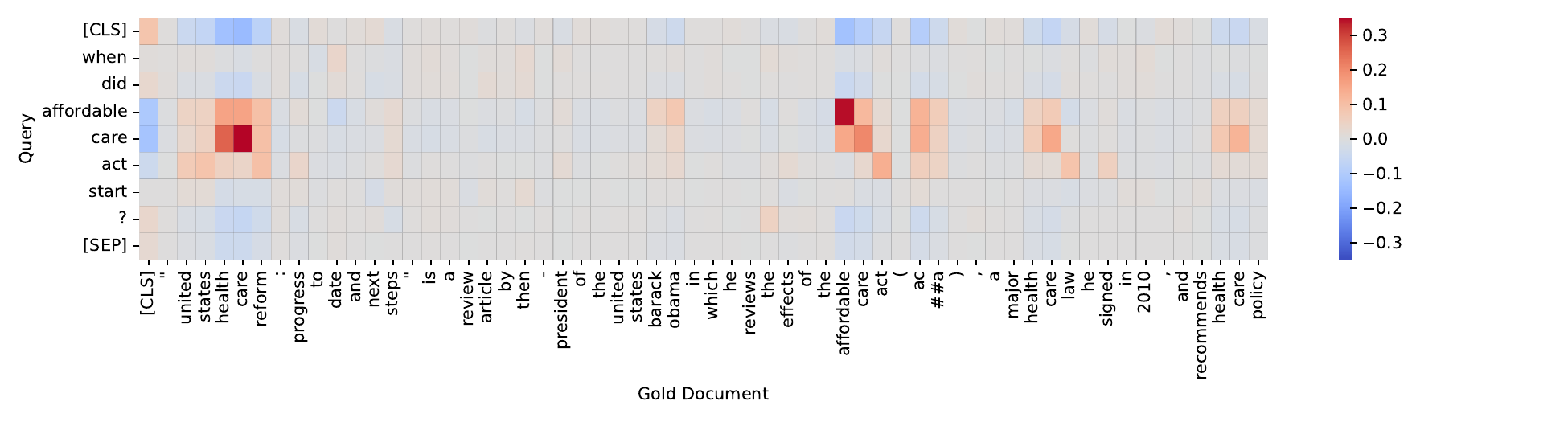}
    \caption{
    Visualization of token-wise effects on retriever scores using DecompX.
    }
    \label{fig:decompx_heatmap_864}
\end{figure*}